\documentclass[conference]{IEEEtran}
\usepackage{amssymb}
\usepackage{amsmath}
\usepackage{graphicx} 
\usepackage{times}
\usepackage{epsfig}
\usepackage{graphicx}
\usepackage{amsmath}
\usepackage{amssymb}
\usepackage{float}
\usepackage{comment}
\usepackage{lipsum}
\usepackage{caption}    
\usepackage{subfig}    
\usepackage[most]{tcolorbox}
\usepackage{hyperref}

\IEEEoverridecommandlockouts  
\title{\LARGE \bf
COCrIP: Compliant OmniCrawler In-pipeline Robot
}

\author{Akash Singh$^{1}$, Enna Sachdeva$^{1}$, Abhishek Sarkar$^{1}$, K.Madhava Krishna$^{1}$% <-this % stops a space
\thanks{*This work was not supported by any organization}% <-this % stops a space
\thanks{$^{1}$All authors are with Robotics Research Center, IIIT-Hyderabad,  Gachibowli-500032, India}%
\thanks{{\tt\small akashvnit2016@gmail.com}}%
\thanks{{\tt\small sachdeva.enna@research.iiit.ac.in}}%
\thanks{{\tt\small abhishek.sarkar@iiit.ac.in}}%
\thanks{{\tt\small mkrishna@iiit.ac.in}}%
}

\begin{document}

\maketitle
\thispagestyle{empty}
\pagestyle{empty}

%%%%%%%%%%%%%%%%%%%%%%%%%%%%%%%%%%%%%%%%%%%%%%%%%%%%%%%%%%%%%%%%%%%%%%%%%%%%%%%%
\begin{abstract}
This paper presents a modular in-pipeline climbing robot with a novel compliant foldable OmniCrawler mechanism. The circular cross-section of the OmniCrawler module enables a holonomic motion to facilitate the alignment of the robot in the direction of bends. Additionally, the crawler mechanism provides a fair amount of traction, even on slippery surfaces. These advantages of crawler modules have been further supplemented by incorporating active compliance in the module itself which helps to negotiate sharp bends in small diameter pipes. The robot has a series of 3 such compliant foldable modules interconnected by the links via passive joints. For the desirable pipe diameter and curvature of the bends, the spring stiffness value for each passive joint is determined by formulating a constrained optimization problem using the quasi-static model of the robot. Moreover, a minimum friction coefficient value between the module-pipe surface which can be vertically climbed by the robot without slipping is estimated. The numerical simulation results have further been validated by experiments on real robot prototype.
\end{abstract}

%%%%%%%%%%%%%%%%%%%%%%%%%%%%%%%%%%%%%%%%%%%%%%%%%%%%%%%%%%%%%%%%%%%%%%%%%%%%%%%%
\section{INTRODUCTION}
In-pipeline climbing robots are highly desirable for non-destructive testing (NDT), inspection and maintenance of complex pipeline networks. Moreover, the widely spread oil and gas pipelines buried beneath under the sea can only be accessed by a robot capable of overcoming bumps, sharp edged joints, valves and bends, while traversing. This has led various researchers and practitioners to design robots for complex small diameter pipe networks, since most of the industrial applications including power plants, boilers etc., and indoor applications deploy pipes of diameter smaller than 100mm. 

A recent state of the art of wheeled, caterpillar, legged, inchworm, screw and Pipe Inspection Gauges (PIG) type pipeline robots is well documented in \cite{c1}. To primarily enhance the traveling performance and speed of locomotion in small diameter pipes, numerous multi-linked wheeled robots have been developed. Dertien.et.al \cite{c2} proposes an omnidirectional wheeled robot for in-pipe inspection (PIRATE) to overcome smooth bends. Hirose  \cite{c3} proposed a `PipeTron' series of multilink articulated snake-like robots with active wheels. PipeTron-I maintains a zig-zag posture to clamp in the pipe, by pulling a pair of steel wires that goes along its backbone and the differential tension in wires provides the twisting motion required to bend the robot.  Further, in PipeTron-VII, twisting motion was realized by the differential speed of each driving wheel. Similarly, a series of Multifunctional Robot for IN-pipe inSPECTion (MRINSPECT)robot  \cite{c4}, \cite{c5},  \cite{c6},  \cite{c10} has been developed for a range of pipe diameters, where MRINSPECT IV  \cite{c6} was specifically designed for pipelines of $\o$ 4 inches. However, the robot fails to realize backward motion in the T junction when its rear module loses contact with the pipe surface. All these articulated robots including \cite{c7}, use passive clamping mechanism, which reduces the number of actuators and therefore, simplifies the control strategy.

\begin{figure}[t]
\centering
\hspace{4cm}
\includegraphics[width=0.4\textwidth,height=0.18\textheight]{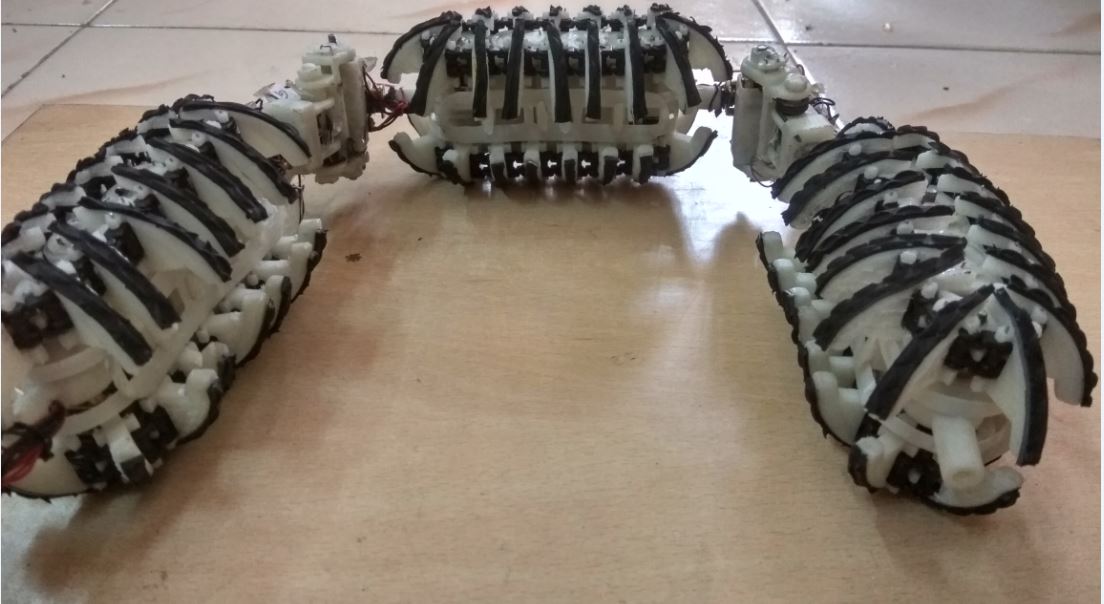}
\caption{Prototype of the Proposed Design }
\label{fig:robot_view} 
\end{figure}

To further simplify the control, a number of screw mechanism based robots have been designed for small diameter pipelines. They have the advantage of achieving efficient helical motion inside pipes with simple transmission mechanism, using a single actuator for driving and rolling motion. Hirose \cite{c8} designed 'Thes-II' robot by interconnecting multiple screw driving modules for \textbf{\O}50mm gas pipelines for long-distance locomotion. Atsushi \cite{c9} designed a screw driving robot using 2 actuators with one being used for driving and rolling motion and the other to select pathways(for steering mode) in branched pipes. However, it is ineffective for autonomous navigation in a known pipe environment due to its unexpected mobility as a result of differential drive.

All the above discussed robots belong to a category of wheeled robots and fail to provide sufficient traction force for climbing on smooth and sticky surfaces. They get stuck on the uneven pipe surface and experience the problem of wheel slip in low friction surface. To tackle these issues, a number of tank-like crawler robots have been developed \cite{c11,c12}. However, due to their inability to realize easy and efficient turning motion and steerability in bends, they suffer from motion singularity at the T-junctions. While Kwon \cite{c11} tried to address this problem with a series of crawler modules, it leads to an increase in the size and weight of the robot. Despite providing greater traction in low friction surfaces,  conventional crawler modules sink into the soft surface while inclined\cite{c13}. Also, no evidence showing the navigation from bigger to smaller diameter pipes, as well as bending in sharp turns, is provided by the previous crawler pipeline robots. This paper seeks to address these problems with the design of a modular Compliant OmniCrawler In-Pipeline climbing robot(COCrIP). The design of the OmniCrawler module, being inspired by \cite{c13}, employs a series of lugs which results in a significant increase in the contact area with the pipe surface. Therefore, the lugs-pipe surface interface provides better traction and the circular cross-section avoids the problem of sinking of modules in the soft surfaces, which makes it robust enough to crawl inside pipes with lower friction coefficient values. The proposed design consists of three such modules interconnected by links with the passive torsion spring joints. The holonomic motion of the robot is achieved by the omnidirectional property of the module, which allows the robot to align itself in the direction of bends. To negotiate sharp turns in small diameter pipes, each OmniCrawler module is further made compliant by incorporating series elastic actuators in it. This is one of the key novelties of this paper. Hence, the modularity of the robot handles pipe diameter variations(lower to higher as well as higher to lower diameters) and the compliance enables it to overcome sharp turns, while exploiting the advantages of OmniCrawler modules.

The organization of the paper is as follows. In Section \ref{Mechanical Design Overview}, a detailed description of the robot mechanism is discussed. Section \ref{in_pipe} discusses the various locomotion modes of the robot to traverse in pipes. Section
\ref{optimization} discusses the optimization framework to estimate an optimal spring stiffness followed by determining the limiting value of friction coefficient between th pipe and the module surface, which the robot can climb vertically with the designed spring values. Furthermore, simulations and experimental results to validate the proof of concept of the mechanical design and mathematical formulation are given in Sections \ref{simulation} and \ref{experiment}, respectively.

\section{Mechanical Design Overview}
\label{Mechanical Design Overview}
The robot has a kinematic chain of 3 compliant foldable OmniCrawler modules with a link connected by a passive joint between two adjacent modules. The detailed description of these parts is given in the subsequent sections. 

\begin{figure}[t]
\centering
%\hspace{-2cm}
\begin{comment}
\subfloat[Exploded side view ]{\includegraphics[width=0.4\textwidth,height=0.18\textheight]{cad_bearing.PNG}}
\end{comment}
\hspace{0cm}
\includegraphics[width=0.5\textwidth,height=0.2\textheight]{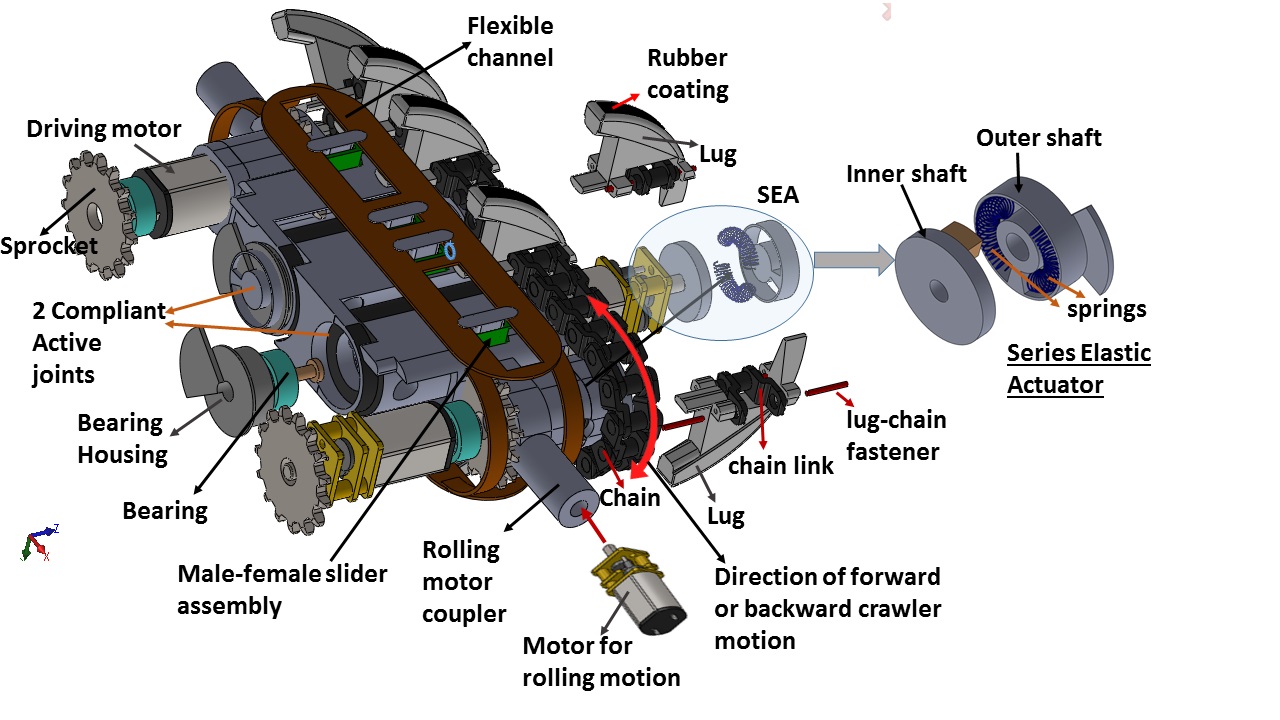}
\caption{Exploded view of the module}
\label{fig:cad model} 
\end{figure}

\begin{figure}[t]
\centering
%\hspace{-2cm}
\begin{comment}
\subfloat[Exploded side view ]{\includegraphics[width=0.4\textwidth,height=0.18\textheight]{cad_bearing.PNG}}
\end{comment}
\hspace{1cm}
\includegraphics[width=0.5\textwidth,height=0.2\textheight]{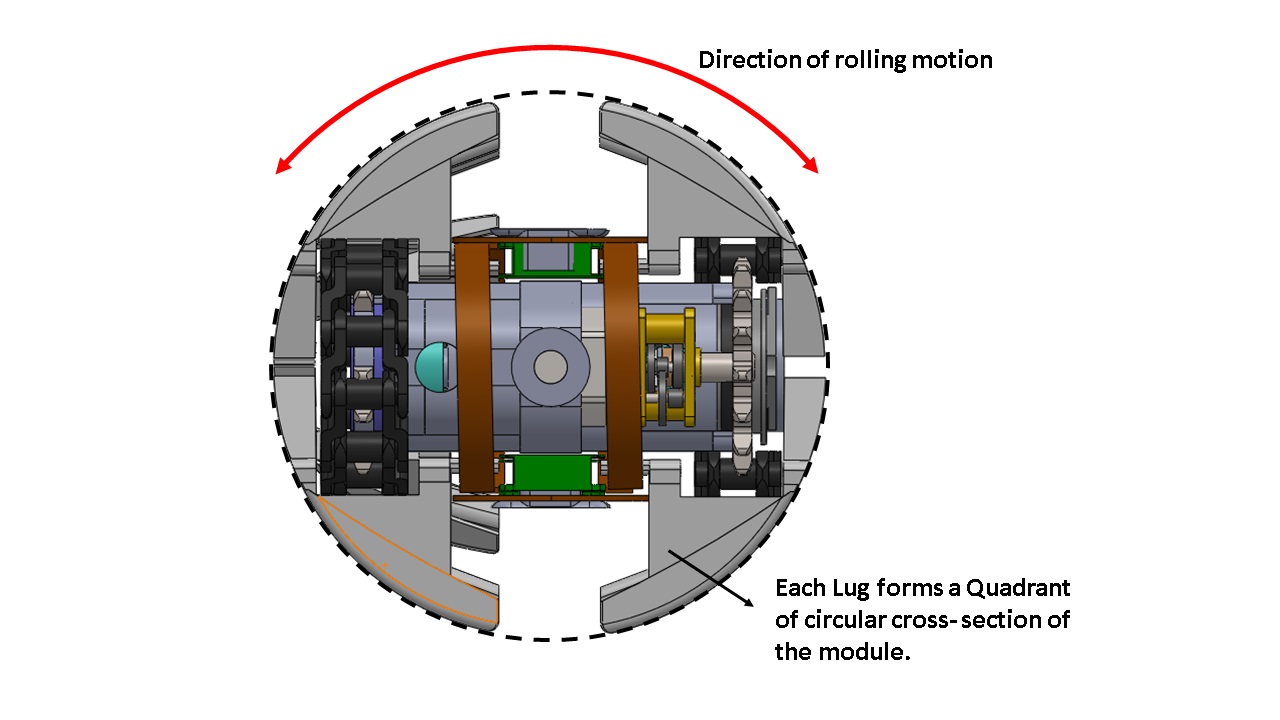}
\caption{Corss-sectional view of module.}
\label{fig:cross_section} 
\end{figure}
 
\begin{table}[h!] 
\caption{Design Parameters of the robot}
\label{table:Design Parameters}
\centering
\begin{tabular}{|l|c|r|}
\hline
\textbf{Quantity} & \textbf{Symbol} & \textbf{Values}  \\
\hline
mass of module & $m_m$ &  0.150kg  \\
\hline
mass of link & $m_l$ &  0.020kg  \\
\hline
length of modules & $l_1$,$l_2$,$l_3$ & 0.14m\\
\hline
Diameter of modules & $d$ & 0.050m  \\
\hline
length of links & $L_1$,$L_2$ & 0.060m  \\

\hline
Diameter of pipe & $D$  & 0.065m to 0.1m  \\
\hline
Driving motors saturation torque & $\tau_{max}$  & 1Nm  \\
\hline
\end{tabular}
\end{table}

\subsection{Foldable OmniCrawler module}

The OmniCrawler module consists of a couple of chain-sprocket power transmission pairs on both sides of the chassis, each of which is driven by a micro motor to provide driving motion in the forward and backward direction. An exploded CAD model view to accompany the following description is shown in Fig. \ref{fig:cad model}. The holonomic motion of the module is characterized by the circular cross-section of the lugs. The design and arrangement of the module components is done such that the diameter of the lugs is minimum based on size constraints of actuators. Two identical longitudinal series of lugs resting on chain links through attachments gives a circular cross-section to the module and the ends of the module attain hemispherical shape, as shown in Fig. \ref{fig:cross_section} . The lugs are coated with a layer of latex rubber, to provide sufficient traction for climbing. The sideways rolling motion of each module is realized with an external actuator connected to the module's chassis with a coupler. The arrangement of the components including chains, lugs, motor mounts, bearings, couplers and chassis are shown in Fig. \ref{fig:cad model}.
%This facilitates the turning of modules in 

\begin{figure}[h!]
\centering
%\hspace{-2cm}
\subfloat[Top view of the straight configuration.]{\includegraphics[width=0.22\textwidth,height=0.10\textheight]{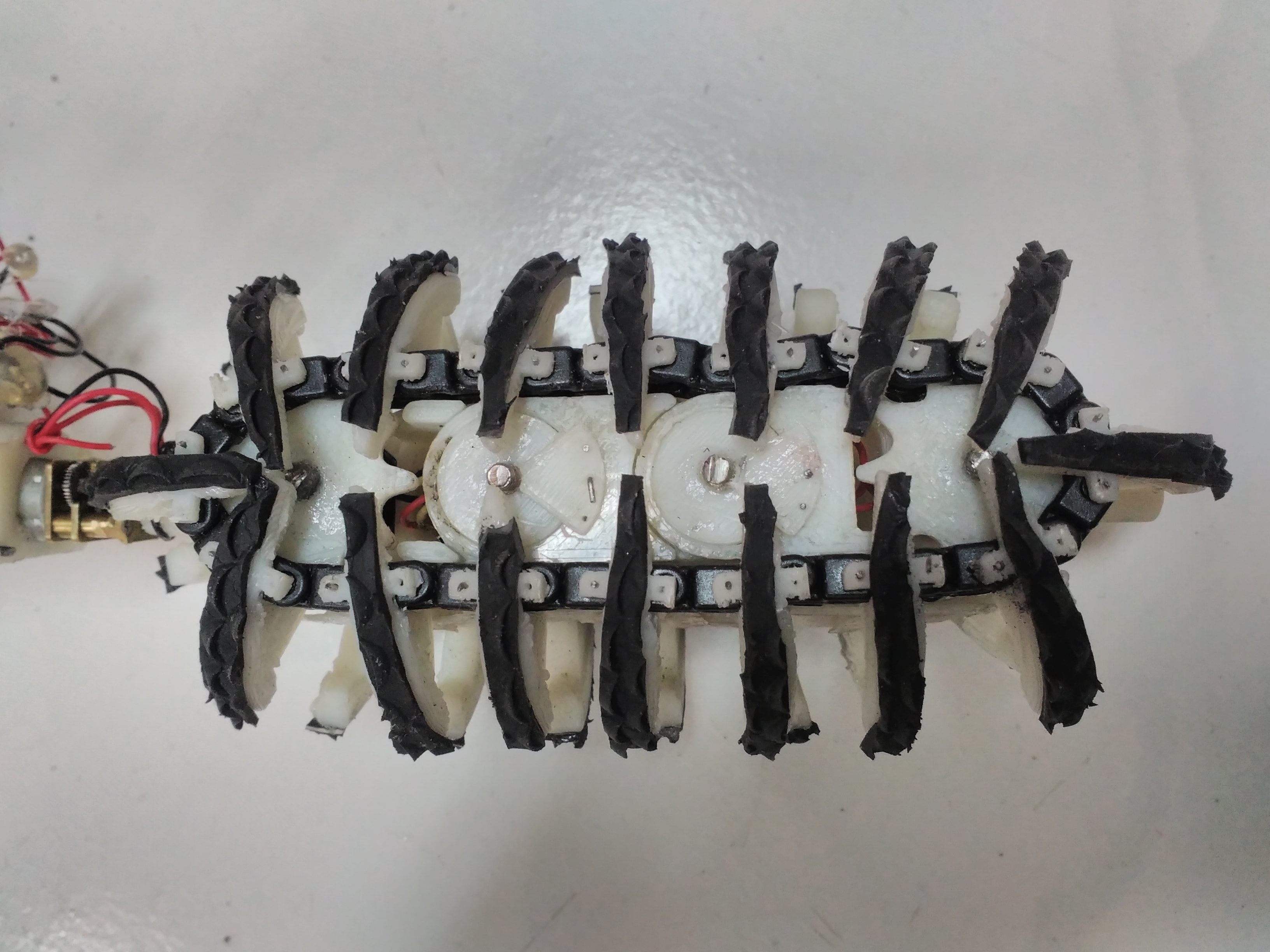}\label{fig:straight_module_top} }
\hspace{0cm}
\subfloat[Side view of the straight configuration.]{\includegraphics[width=0.22\textwidth,height=0.10\textheight]{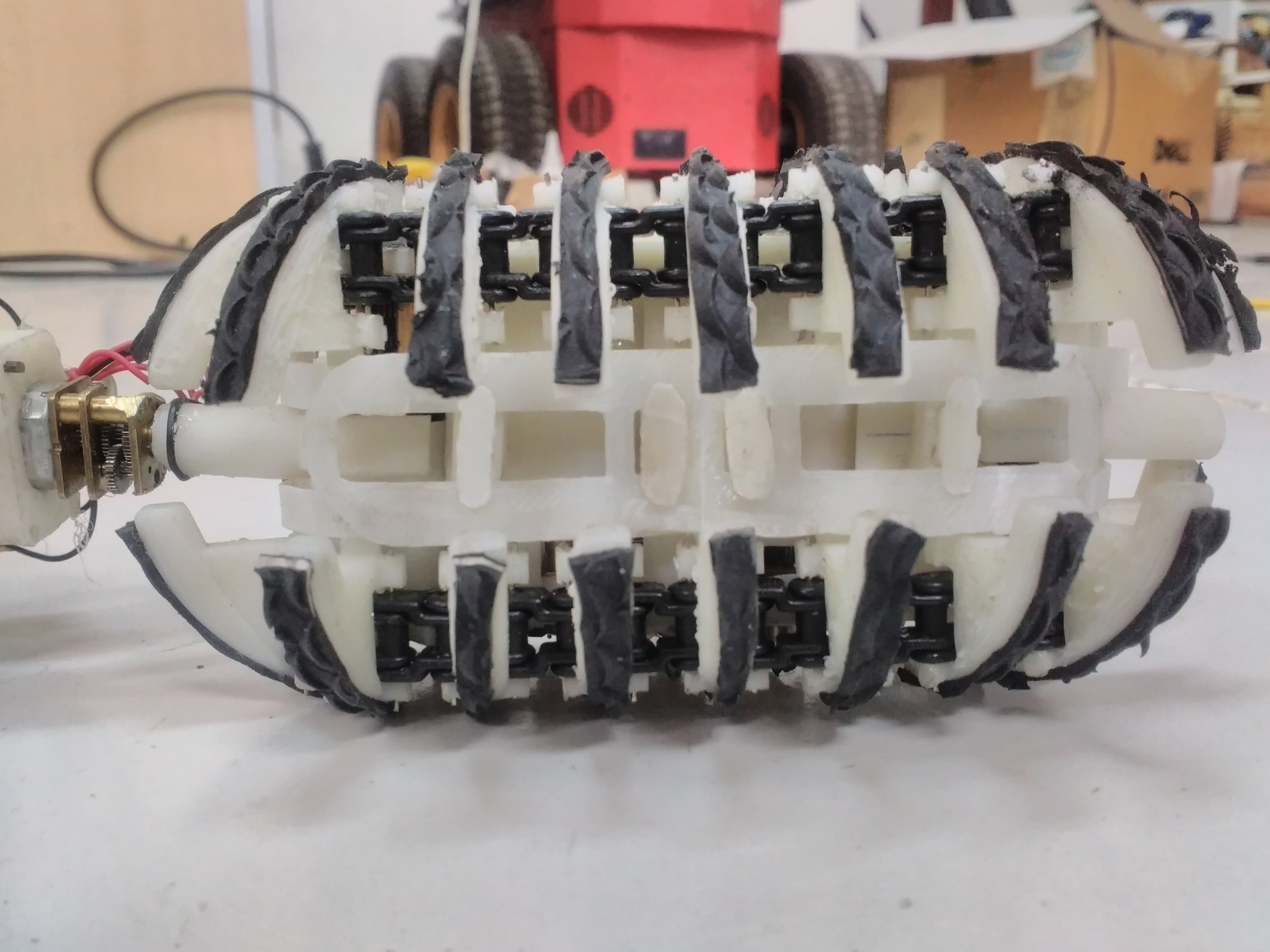}\label{fig:straight_module_side} }\\
\subfloat[Top view of the bend configuration]{\includegraphics[width=0.22\textwidth,height=0.10\textheight]{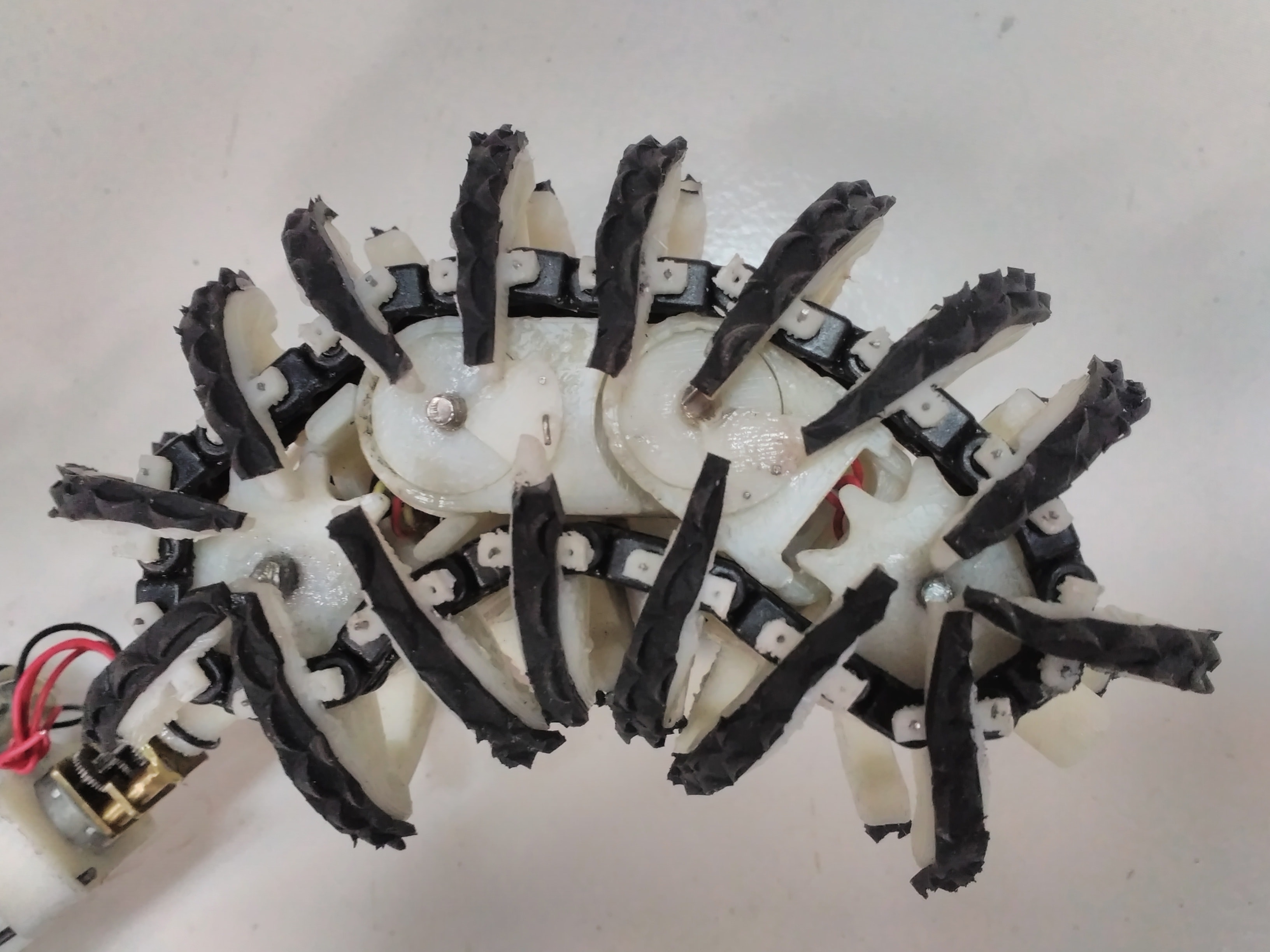}\label{fig:bend_module_top} }
\hspace{0cm}
\subfloat[Side view of the Bend configuration]{\includegraphics[width=0.22\textwidth,height=0.10\textheight]{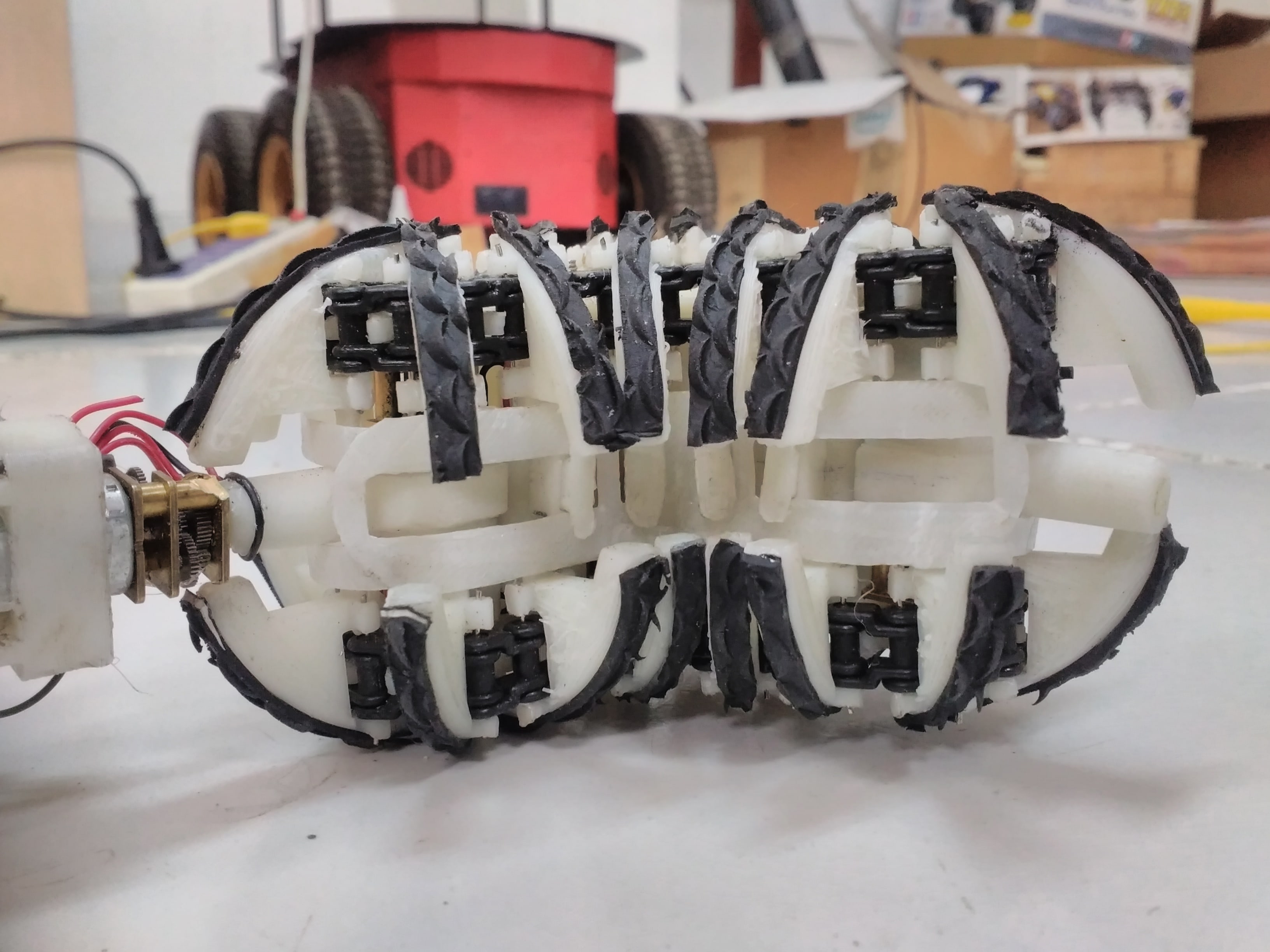}\label{fig:bend_module_side} }
\caption{Module in straight and bend configuration with the active compliant hinge joints.}
\label{fig:modules_orientation} 
\end{figure}

The mechanism to achieve compliance of the OmniCrawler module is realized by incorporating 2 active hinge joints, in the support structure(chassis) of the module. Each such revolute joint is actuated by a geared motor with its shafts connected in series with an assembly of dual shafts(inner and outer shaft) and linear springs. The inner shaft is connected to the motor shaft and the outer shaft to the hinge joint. This arrangement of geared motor with the linear springs acts as a series elastic actuator(SEA) \cite{c16} and is being used to achieve module compliance. It further filters out vibrations while overcoming jagged terrains during forward propulsion of the robot and thereby, protects the gears of the joint motors from getting damaged. To ensure that the chains as well as lugs align themselves along the bent chassis, a flexible channel goes over the entire module's body and provides a passage to the lugs-chain assembly. This is achieved with the aid of channel-lug-chain attachments. Furthermore, a series of coupled male-female slider assembly slides through the flexible channel cavity and enables the lug-channel assembly to maintain a constant height above the chassis, throughout. This slider assembly also acts as stopper and restricts compliant hinge joint rotations beyond its joint angle limits. This is clearly depicted in Fig.\ref{fig:bend_module_side}.

\subsection{Link design}

The link assembly is connected to each of the adjoining modules with a compliant passive joint. The design of the link is determined by the desired pipe diameter as well bend curvature. To overcome sharp $90^\circ$ bend in a \textbf{\O}75mm pipe, the link design is optimized to incorporate rolling motor clamps of the adjoining modules, as well as a pair of torsion springs corresponding to each compliant passive joint. 

\section{In-pipe locomotion}
\label{in_pipe}
\begin{comment}
In order to get sufficient traction force to climb inside pipes, an optimal torsion spring stiffness to clamp the modules with the pipe walls has been obtained in section III.
\end{comment}

\subsection{Locomotion mode in straight pipes}
\begin{figure}[h!]
\centering
%\hspace{-2cm}
\subfloat[Rolling Motion about the axis of the pipe ]{\includegraphics[width=0.15\textwidth,height=0.18\textheight,angle=-0] {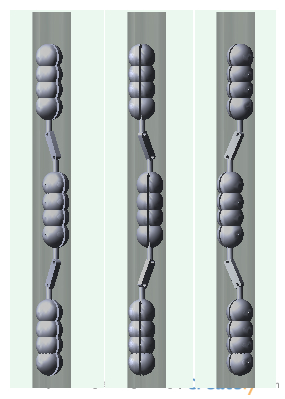}\label{fig:rolling}} 
\hspace{1cm}
\subfloat[Maximum energy posture(left) and minimal energy posture(right) for turning ]{\includegraphics[width=0.25\textwidth,height=0.16\textheight, angle=-0]{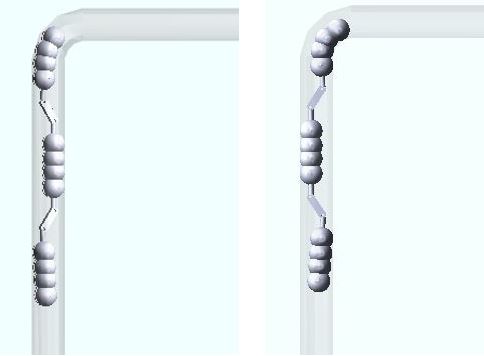}\label{fig:energy}}

\caption{Robot configuration in straight pipes and bends}
\end{figure}

In straight pipes, all 3 modules are aligned in-line with the pipe and driven synchronously to propagate the robot in forward/backward direction. The preloaded torsion spring joints provide the necessary clamping force to overcome robot's own body weight and facilitate slip free driving motion. The alignment of the robot in straight configuration is shown in Fig.\ref{fig:rolling} and \ref{fig:vert}.

\subsection{Locomotion mode in bend pipes}
The robot rolls about the axis of the pipe, to align its orientation to steer in bends. All 3 module rotates synchronously, such that they remain in-line with the pipe and their axes continue to remain in the same plane, as shown in Fig.\ref{fig:rolling}. In order to negotiate bends in a minimum energy posture, the compliant passive joints need to apply minimal torque during bending, as shown in Fig.\ref{fig:energy}. Therefore, the rolling motion is executed to attain the minimal energy posture. 
\begin{table}[t] 
\hspace{-2cm}
\centering
\caption{Nomenclature for model description}
\begin{tabular}{|p{2cm}|p{5cm}|}
\hline
\textbf{Symbols} & \textbf{Quantity}  \\
\hline
$k_1,k_2,k_3,k_4$ & torsion spring constant of 4 passive joints \\

\hline
$ij$ & represents $j_{th}$ sub-module of $i_{th}$ module, \\

\hline
$wm_{ij}$ & weight of $ij_{th}$ sub-module \\

\hline
$d$  & Diameter of the module \\

\hline
$l_{ij}$  &  length of sub-module \\

\hline
$F_{ij}$  &  Friction force of $ij^{th}$ sub-module\\

\hline
$N_{ij}$ & Normal force acting on $ij^{th}$ sub-module\\

\hline
$\theta_{ij}$  &  angle of $ij^{th}$ sub-module with global x(horizontal) axis\\

\hline
$J_{ij}$  &  represents joint between $i^{th}$ and $j^{th}$ sub-module \\

\hline
$wl_k$ & weight of $k^{th}$ link ( $k^{th}$ link connects  $k^{th}$ module with  $k+1^{th}$ module\\

\hline
$L_k $ &  length of  $k^{th}$ link\\

\hline

$\theta_k$  &  angle of $k^{th}$ link with the horizontal axis \\
\hline

$D$  &  Diameter of the pipe (represented as \textbf{\O}) \\

\hline
$\mu$  & coefficient of friction \\

\hline
$f_{x}$  & force acting in $x$ direction \\

\hline
$f_{y}$  & force acting in $y$ direction \\

\hline
$M_J$  &  Moment acting on joint J \\

\hline
$\tau_k$  & torque \\

\hline

\end{tabular}
\label{table:Nomenclature}
\end{table}

\section{Optimization Framework}
\label{optimization}

An optimal spring stiffness estimation is critical for providing sufficient clamping force to climb the robot in vertical straight and bend pipes. Lower values of spring stiffness result in insufficient traction and lead to slippage, whereas higher values result in a larger moment at these joints. 
\begin{comment}
Fig.\ref{eqn:non_opt_spring} illustrates the misalignment of modules leading to reduction of point of contact with the pipe surface, as a consequence of too lower and higher spring stiffness, respectively.  
\begin{figure}[h!]
\centering
\hspace{-0cm}
\subfloat[With lower values of spring stiffness at $J_1$ and $J_4$ ]{\includegraphics[width=0.12\textwidth,height=0.14\textheight,angle=-0] {vert_spr_2_snap.PNG}\label{fig:L_turn_2}} 
\hspace{1cm}
\subfloat[With higher values of spring stiffness at $J_1$ and $J_4$  ]{\includegraphics[width=0.13\textwidth,height=0.14\textheight, angle=-0]{vert_spr_1_snap.PNG}}

\caption{Robot configuration with non-optimal spring stiffness values}
\label{eqn:non_opt_spring}
\end{figure}
\end{comment}

\subsection{Determining Optimal Spring Stiffness}
The estimation of optimal spring stiffness is formulated as an optimization problem with an objective to minimize the sum of torsion spring joint moments, which ensures that the spring is not too stiff.   

\begin{align}
\begin{split}
min \sum_{j=1}^{4}|\tau_j|
\end{split}
\end{align}

This function being linear and convex, is guaranteed to converge to a global optima. The constraints posed by the geometry, model as well as motion of the robot to perform this optimization, is discussed in the subsequent sections.
\subsubsection{No slip constraint}
In order to avoid slippage at robot-pipe surface interface, friction force which directly relates with the wheel torque must satisfy the following constraint.
\begin{align}
\begin{split}
F_{ij}=F_{static} \le \mu N_{ij}, \forall i,j \in \{1,2,3\}
\end{split}
\label{eqn:noslip}
\end{align}
Moreover, the maximum traction force is constrained by the driving motor torque limits.

\begin{align}
\begin{split}
 F \le \frac{2\tau_{max}}{(d/2)}
\end{split}
\label{eqn:motortorque}
\end{align}

To determine these unknown parameters $(N_{ij}, F_{ij}, \tau_{ij})$, we need to formulate a model of the robot to associate all the forces and torques with the spring joint moments. As the robot crawls at a speed of 0.5 m/sec and at such low speed, the motion of a robot is dominated by the surface forces rather than dynamic and inertial effects \cite{c14}, the slow motion crawling behavior can be well captured by the quasi-static model of the robot, which is discussed in the following section.

\subsubsection{Quasi-static analysis and  Kinematic constraints}
For the mathematical analysis, each submodule is modeled as a symmetric Omniball, as proposed by \cite{c13}.

\begin{itemize}
\item[(i)]\textit{In straight configuration}\\
With the modules, aligned in-line with the straight pipes, as shown in Fig.\ref{fig:vert}, the joint angles $\theta_1,\theta_2$ are geometrically determined as following.

\begin{figure}[t]
\centering
\includegraphics[height=0.2
\textheight,width=0.5\textwidth, angle=0.4]{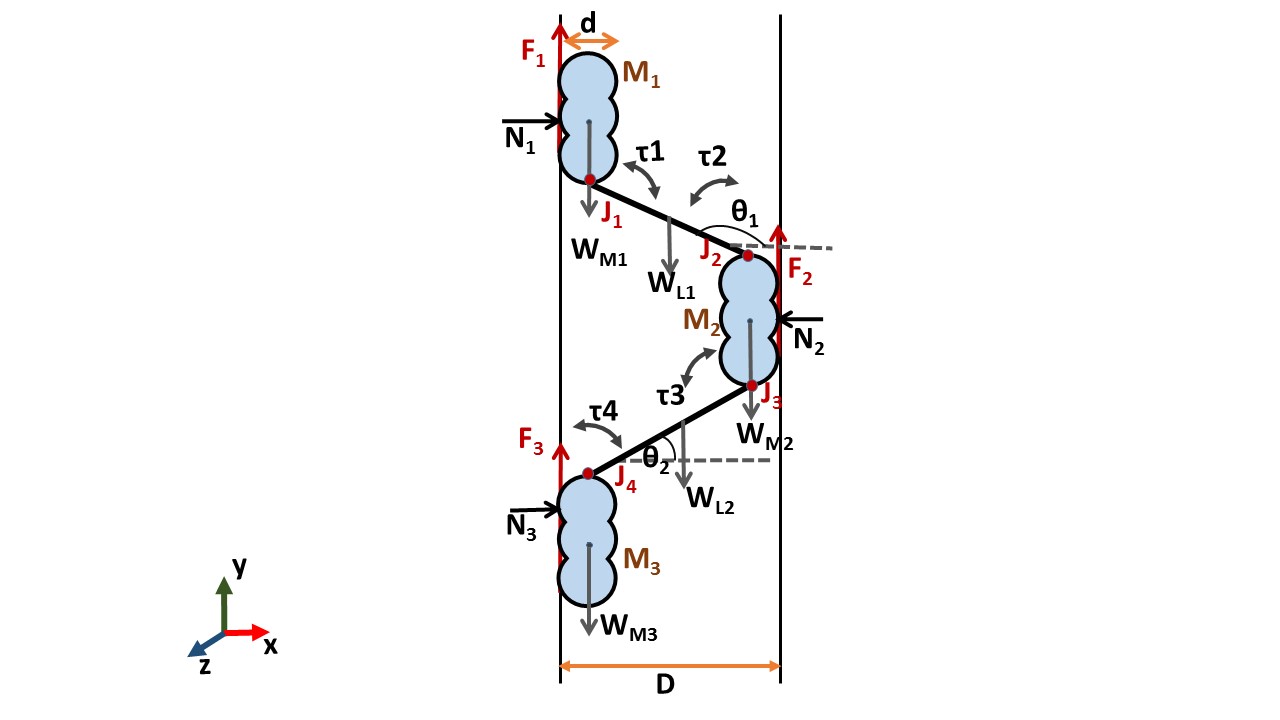}
\caption{Quasi-static configuration in vertical straight pipe}
\label{fig:vert}
\end{figure}

\begin{figure}
\centering
\hspace{-0cm}
\includegraphics[height=0.22
\textheight,width=0.5\textwidth]{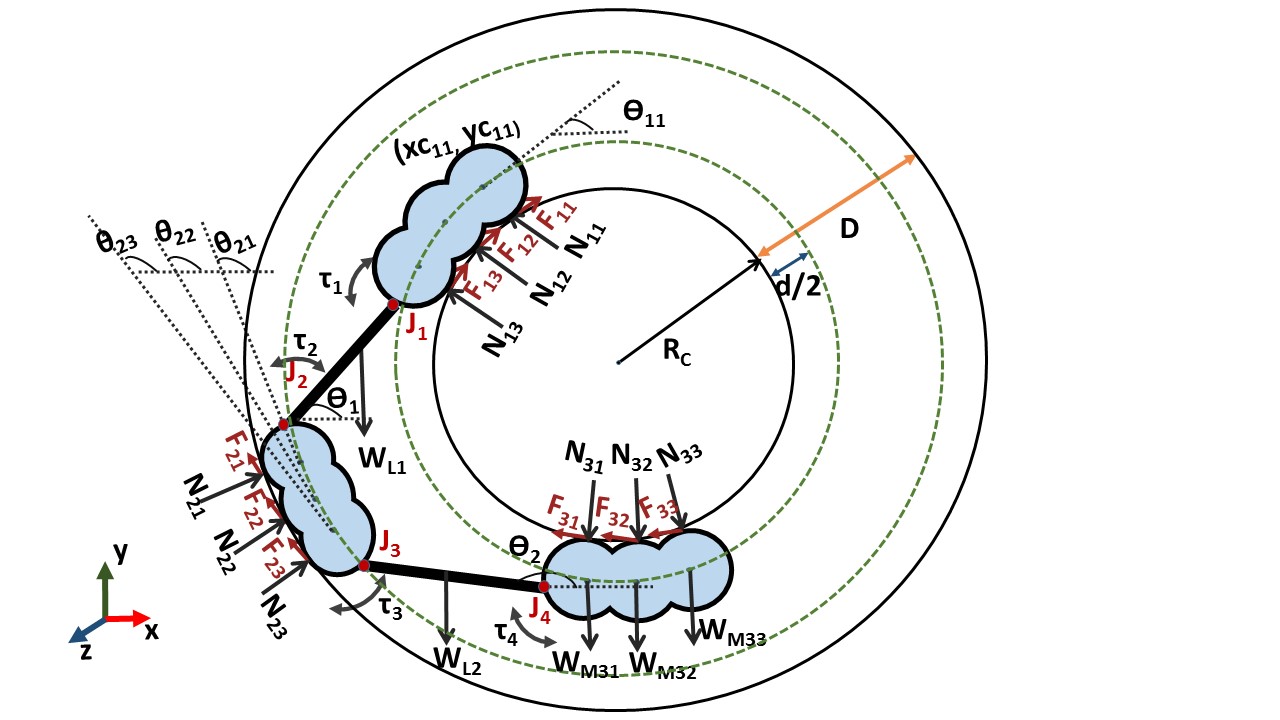}
\caption{Quasi-static configuration in vertical bend pipes}
\label{fig:bend}
\end{figure}

\begin{align}
\begin{split}
&\theta_1=\pi- cos^{-1} (\frac{D-d}{L_1}) \\ 
&\theta_2=cos^{-1}  (\frac{D-d}{L_2})  
\end{split}
\label{eqn:vert_theta}
\end{align}

With the posture parameters($\theta_1,\theta_2$), the quasi static model can be obtained by balancing the forces(equation.\ref{eqn:f_x},\ref{eqn:f_y}) and joint moments at $J_1,J_2,J_3$ and $J_4$ (equations.\ref{eqn:MJ1}-\ref{eqn:MJ4}), as follows.

\begin{align}
\begin{split}
\hspace{-2cm}\Sigma f_x=0, \quad \quad \quad \quad N_1  - N_2 + N_3= 0
\end{split}
\label{eqn:f_x}
\end{align}

\begin{align}
\begin{split}
\Sigma f_y=0, \quad \quad \quad \quad     & F_1 + F_2 + F_3-wm_1-wm_2\\
&-wm_3-wl_1-wl_2=0
\end{split}
\label{eqn:f_y}
\end{align}

\begin{align}
\begin{split}
\hspace{-1cm}\Sigma M_{J_1}=0, \quad \quad \quad F_1d/2+N_1l_1/2-\tau_1=0
\end{split}
\label{eqn:MJ1}
\end{align}

\begin{equation} 
\begin{split} 
\Sigma M_{J_2}=0, \quad  \quad \quad 
&F_1L_1\cos\theta_1+N_1L_1\sin\theta_1-\\
&wm_1L_1\cos\theta1-
wl_1L_1/2\cos\theta_1+\\
&\tau_1-\tau_2=0
\end{split} 
\label{eqn:MJ2}
\end{equation}

\begin{equation} 
\begin{split} 
\Sigma M_{J_3}=0, \quad  \quad \quad 
&-F_2d/2+N_1l_1-N_1l_2/2+\\
&\tau_2-\tau_3=0
\end{split} 
\label{eqn:MJ3}
\end{equation} 

\begin{equation} 
\begin{split} 
\Sigma M_{J_4}=0, \quad  \quad  \quad
&-F_1L_2\cos\theta_2+N_1L_2\sin\theta_2-\\
& F_2L_2\cos\theta_2-N_2L_2\sin\theta_2+\\ 
&(wm_2+wm_1+wl_1)L_2\cos\theta_2+\\ &wl_2(L_2/2)\cos(\theta_2)+ \\
&\tau_3-\tau_4=0; \\
& N_3l_3/2-F_3d/2-\tau_4=0
\end{split} 
\label{eqn:MJ4}
\end{equation} 

\item[(ii)]\textit{In bend configuration}\\
The desired posture of the folding Omni-Crawler modules in the bent configuration should comply with the geometry of curvature of the bend pipes,as shown in Fig.\ref{fig:bend}. The joint angles of each submodule corresponding to the desired posture can be geometrically determined, subject to the constraint that the center of all sub-modules must trace a circle of radius $R_{cin}$ or 
$R_{cout}$. 

\begin{align}
\label{eqn:eqlabel}
\begin{split}
&R_{cin}=R_c+d/2 \\
&R_{cout}=R_c+D-d/2
\end{split}
\end{align}

\begin{align}
\begin{split}
&(xc_{ij}-xc)^2+(yc_{ij}-yc)^2=R^2  \quad \forall i,j\\ 
& R=R_{cin},  \quad \quad  i \in \{1,3\},j \in \{1,2,3\}\\    
& R=R_{cout}, \quad \quad i \in \{2\},j \in \{1,2,3\}    
\end{split}
\end{align}

where,\\ 
$R_c$: radius of curvature of the pipe\\
$(xc_{ij},yc_{ij})$ : coordinates of center of $ij_{th}$ submodule, as shown in Fig.\ref{fig:bend}\\

After determining the center of each submodule, their joint angles can be be estimated as below.

\begin{align}
\begin{split}
&\theta_{ij}=tan^{-1}( \frac{yc_{ij}-yc}{xc_{ij}-x_c})-\pi/2\\
& \quad \quad \forall i\in  \{1,2,3\}
\end{split}
\end{align}

In a similar way, passive joint angle are determined,

\begin{align}
\begin{split}
&\theta_{k}=tan^{-1}( \frac{yc_{k}-yc}{xc_{k}-xc})-\pi/2, \\
&\quad \forall k\in  \{1,2\}
\end{split}
\end{align}       

$(xc_{k},yc_{k})$ : coordinates of center of mass of $k_{th}$ link

\begin{comment}
\begin{align}
\hspace*{0.5cm}  \Sigma F_X=0  \quad   \quad \quad   \quad   N_1  - N_2 + N_3= 0 
\end{align}

\begin{align}
\hspace*{0.5cm}  \Sigma F_Y=0 \quad   \quad \quad   \quad  &F_1 + F_2 + F_3-wm_1-wm_2\\
														   &-wm_3-wl_1-wl_2=0  
\end{align}

\begin{align}
\hspace*{0.5cm} \Sigma M_A=0       \quad   \quad    \quad     \quad  \quad \quad  &F_1 \frac{d}{2}+N_1\frac{l_1}{2}-\tau_1=0
\end{align}

\begin{align}
\begin{split}
\hspace*{0.8cm}  \Sigma M_B=0    \quad     \quad &-(F_1-wm_1)(L_1\cos{\hat\theta_1})-N_1(L_1\sin{\hat\theta_1})\\
& +wl_1(\frac{L_1}{2} \cos{\hat\theta_1})+\tau_2-\tau_1=0
\end{split}
\end{align}

\begin{align}
\begin{split}
&\hspace*{0.8cm}  \Sigma M_C=0     \quad     \quad F_2 \frac{d}{2} - N_1l_2+N_2\frac{l_2}{2}+\tau_3-\tau_2=0
\end{split}
\end{align}

\begin{align}
\begin{split}
\Sigma M_D=0     \quad     \quad    \quad     \quad   N_3\frac{l_3}{2}-F_3\frac{d}{2}-\tau_4=0
\end{split}
\end{align}
\end{comment}
Quasi-static equations are given by-

\begin{align}
\begin{split}
\hspace{-0.5cm}\Sigma f_x=0, \quad \quad  
& \Sigma F_{ij}\cos(\theta_{ij})-\Sigma N_{ij}\sin(\theta_{ij})=0  
\end{split}
\end{align}

\begin{align}
\begin{split}
\hspace{-1.5cm}\Sigma f_y=0, \quad  \quad  
&\Sigma F_{ij}\sin(\theta_{ij})+\Sigma N_{ij}\cos(\theta_{ij})-\\
&\Sigma wm_{ij}-\Sigma wl_{k}=0 
\end{split}
\end{align}

\begin{align}
\begin{split}
\hspace{-0.33cm}\Sigma M_{J_{12}}=0, \quad \quad
& F_{11}\frac{d}{2}+N_{11}\frac{l_{11}}{2}-\frac{w_{11}}{2}l_{11}\cos\theta_{11}-\\ 
&\tau_{11}=0
\end{split}
\end{align}

\begin{comment}
\begin{align}
\begin{split}
\Sigma M_{J_{12}}=0, \quad
& -F_{11}l_{12}\sin(\theta_{12}-\theta_{11})+\\ & N_{12}l_{12}\cos(\theta_{12}-\theta_{11})+ F_{12}\frac{d}{2}+N_{12}\frac{l_{12}}{2}-\\
&(wm_{11}+\frac{wm_{12}}{2})l_{12}\cos(\theta_{12})+\\
&\tau_{11}-\tau_{12}=0
\end{split}
\end{align}

\begin{align}
\begin{split}
\hspace{-0.5cm}\Sigma M_{J_{13}}=0, \quad  
-& F_{11}l_{13}\sin(\theta_{13}-\theta_{11})+\\
&N_{11}l_{13}\cos(\theta_{13}-\theta_{12})-\\
& F_{12}l_{13}\sin(\theta_{13}-\theta_{11})+\\
&N_{12}l_{13}\cos(\theta_{13}-\theta_{12})+\\ 
& F_{13}\frac{d}{2}+N_{13}\frac{l_{13}}{2}-(wm_{11}+wm_{12}+\\
& wm_{13}/2)l_{13}cos\theta_{13}+\tau_{12}-\tau_1=0\\ 
\end{split}
\end{align}
\end{comment}

A similar set of equations are derived by balancing the moment at each joint. The detailed derivation is uploaded at \url{http://robotics.iiit.ac.in/Archives/pipe_climbing_robot_quasistatic.pdf}.\\
These equations form the equality constraints and are represented as $Ax=b$, where
$x$ is a vector of variables ($x=[(F_{ij})^T,(N_{ij})^T,(\tau_{ij})^T]^T$).
 
Therefore, the optimization can be formulated as 
\begin{align}
\begin{split}
&\min_{\tau_j} \sum_{j=1}^{4}|\tau_j|\\
\text{subject to} \quad &Ax=b,\\ 
                        &F_{ij} \le \mu N_{ij}, \quad \forall i,j \in \{1,2,3\}
\end{split}
\label{eqn:spring_opti}
\end{align}

\end{itemize}

\subsection{Determining the range of friction coefficients for no-slip climbing}
In an in-pipe environment, it is quite difficult to determine an exact value of $\mu$ for the robot-pipe surface interface. Moreover, the friction changes dynamically along the pipe networks.Therefore, it is required to analyze the climbing ability of the robot inside pipes with different friction coefficients. This is achieved by finding a minimum $\mu$ of the surface, on which this robot can climb with the spring stiffness values obtained from equation (\ref{eqn:spring_opti}), without slipping. For a given $\mu$, no-slip condition is achieved by maximizing the traction force while climbing. However, this traction force must neither exceed the maximum force that a terrain can bear nor the motor saturation torques, as stated in equations \ref{eqn:noslip},\ref{eqn:motortorque}. Therefore, the objective of slippage avoidance can be achieved by minimizing the maximum ratio of traction force ($F_{ij}$) to the normal force ($N_{ij}$) \cite{c15} at the point of contact of $ij^{th}$ submodule with the surface of the pipe.
\begin{comment}
An optimization criteria for maximizing traction at the robot-pipe surface interface is based on the observation that the force that a terrain can bear increases with increasing normal forces. Therefore, to avoid terrain failure and wheel slip, the control algorithm should minimize the maximum ratio of traction force to normal force. In other words, 
\end{comment}
\begin{align*}
\begin{split}
R = \max \{F_{ij}/N_{ij}\}
\end{split}
\end{align*}

By minimizing the function $R$, it is ensured that the required minimum traction force per unit normal force is being applied to maintain static stability, in a particular configuration. However, being a non-linear function, the optimization process may get stuck in local optima and strictly depends on a strong initial guess to converge to a value closer to global optima. Therefore, the objective is modified as maximizing the sum of Normal forces acting on all submodules as it eventually minimizes the cost function $R$. 
\begin{align}
\begin{split}
\max_{F{ij}/N{ij}}
 &\sum_{i,j}|N_{ij}|^2\\
\text{subject to,} \quad &Ax=b,\\ 
&F_{ij} \le N_{ij}, \quad \forall i,j \in \{1,2,3\}
\end{split}
\label{eqn:fric_opti}
\end{align}

Since, this optimization is carried out with the stiffness values$(k_1,k_2,k_3,k_4)$ determined from solution to equation \ref{eqn:spring_opti}, the variables here include only $F$ and $N$($\therefore x=[[F_{ij}]^T,[N_{ij}]^T]^T$). 

$\therefore$\hspace{2cm} $\mu_{lim}=\max(F_{ij}/N_{ij})$

\begin{figure*}[t!]  
\centering
\hspace{-1cm}
%\subfloat[Joint moments values versus $\phi$.]%{\includegraphics[width=0.24\textwidth,height=0.15\textheight]{joint_moment.png}\label{fig:joint_moments}}
%\hspace{0cm}
\subfloat[Spring stiffness values versus $\phi$]{\includegraphics[width=0.24\textwidth,height=0.15\textheight]{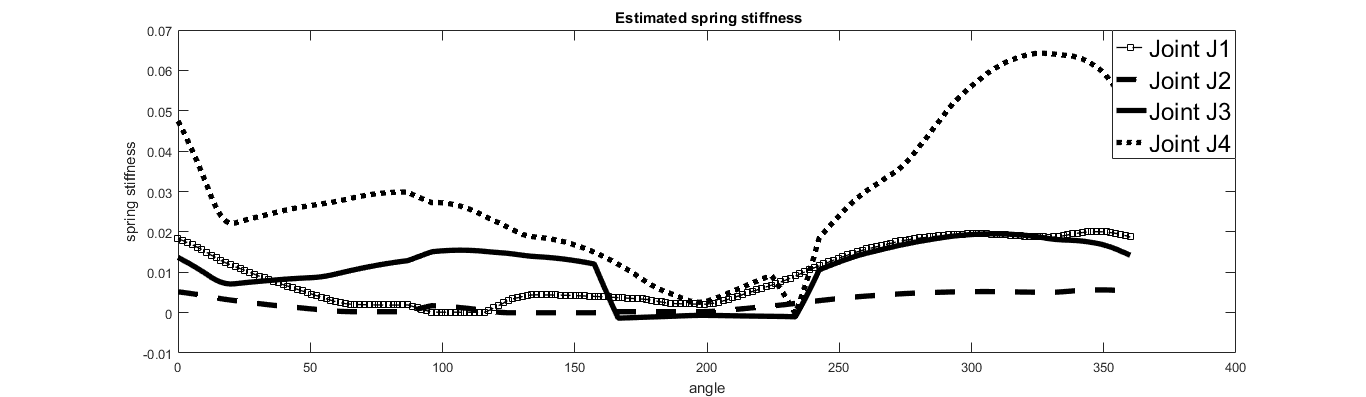}
\label{fig:spring_stiff}}
\hspace{0cm}
\subfloat[Spring stiffness corresponding to $\mu$ and $\mu_{lim}$ ]{\includegraphics[width=0.23\textwidth,height=0.15\textheight]{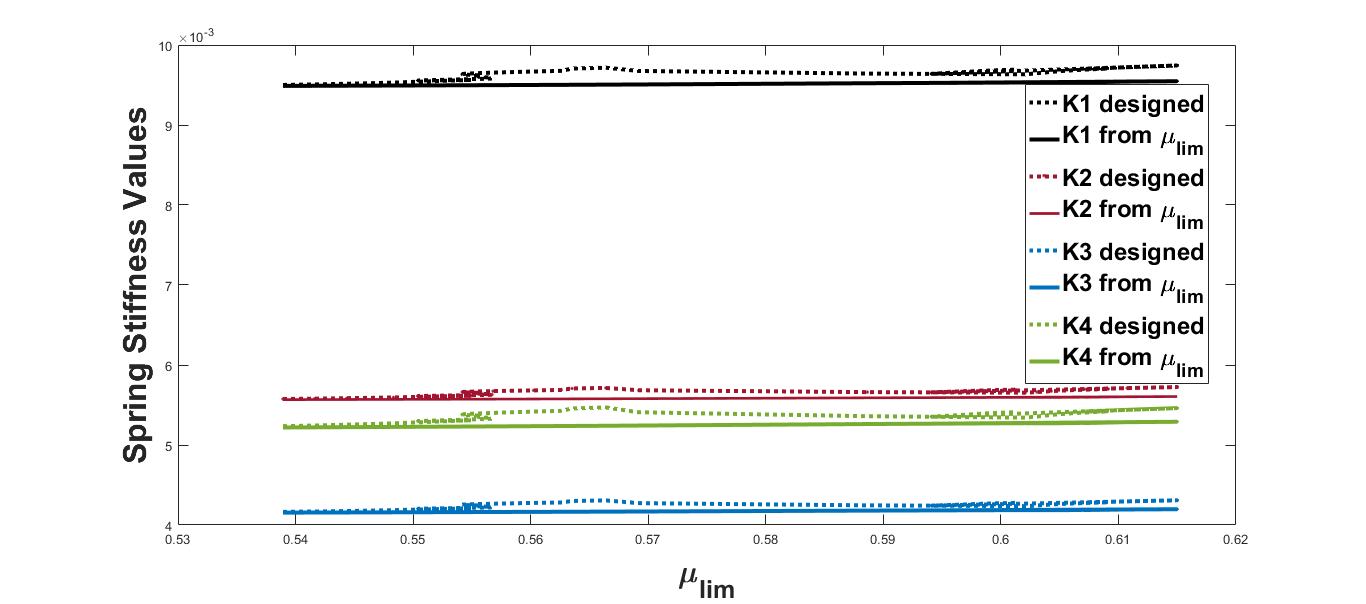}\label{fig:mu_lim}}
\hspace{0cm}
\subfloat[$\mu$ versus $\mu_{lim}$]{\includegraphics[width=0.23\textwidth,height=0.15\textheight]{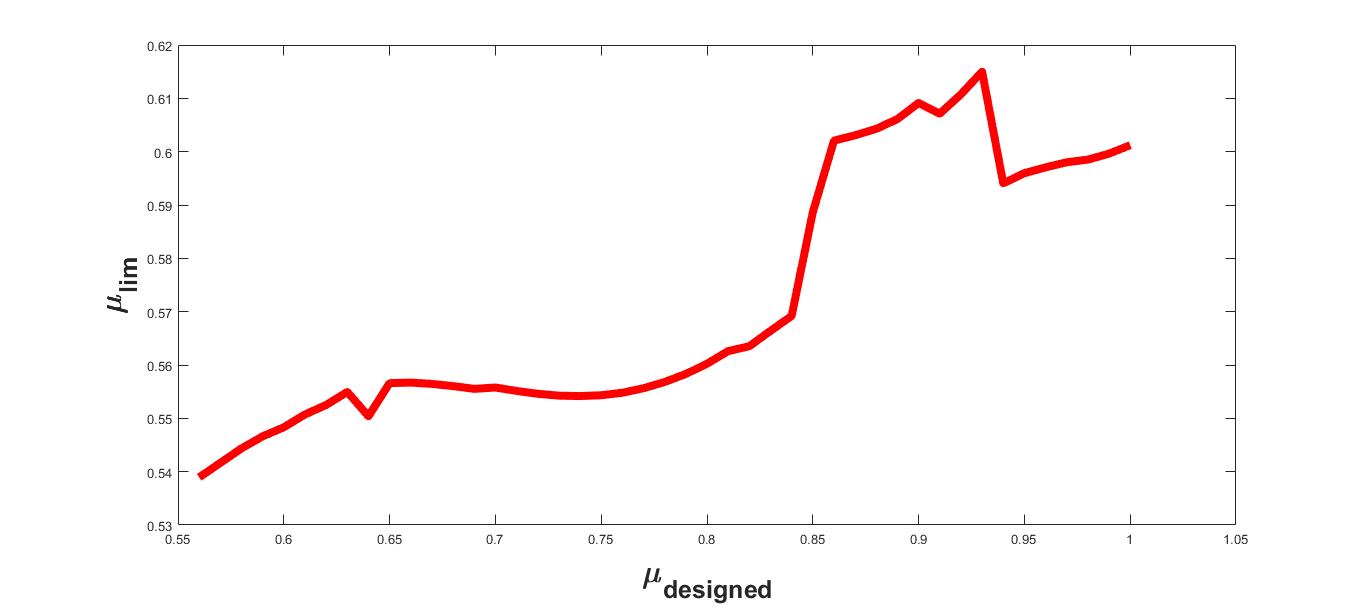}\label{fig:mu_vs_mu_lim}}
\caption[Optional caption for list of figures 5-8] {Plots obtained from optimization at bends.}
\label{fig:matlab_plots}
\end{figure*}

\section{Simulation Results}
\label{simulation}
To validate the proof of concept of the design as well as optimize the design parameters for a desirable pipe environment, simulations were carried out in ADAMS MSC, a multi-body dynamics simulator, with a lumped model of the robot. With the design parameters listed in Table\ref{table:Design Parameters}, the formulated constrained Optimization(eqn.\ref{eqn:spring_opti}) yields a minimal set of passive compliant joint torques at $J_1, J_2, J_3$ and $J_4$, to statically balance the robot, which is further used to obtain the stiffness values, by assuming the linearity of the springs.

\begin{figure*}[htp!]   
\centering

\subfloat[Simulation in $90^\circ$ bent elbow]{\includegraphics[width=1\textwidth,height=0.14\textheight]{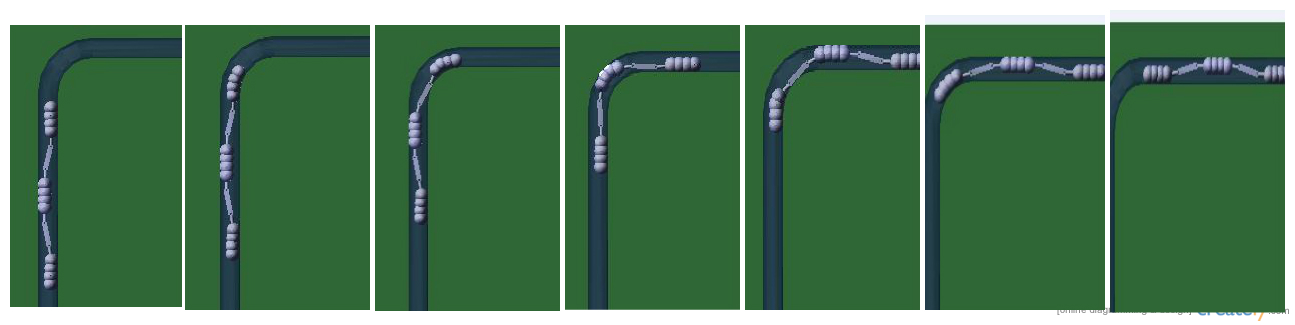}\label{fig:adams_90}} 

\subfloat[Experiment in $90^\circ$ bent elbow]{\includegraphics[width=1\textwidth,height=0.14\textheight]{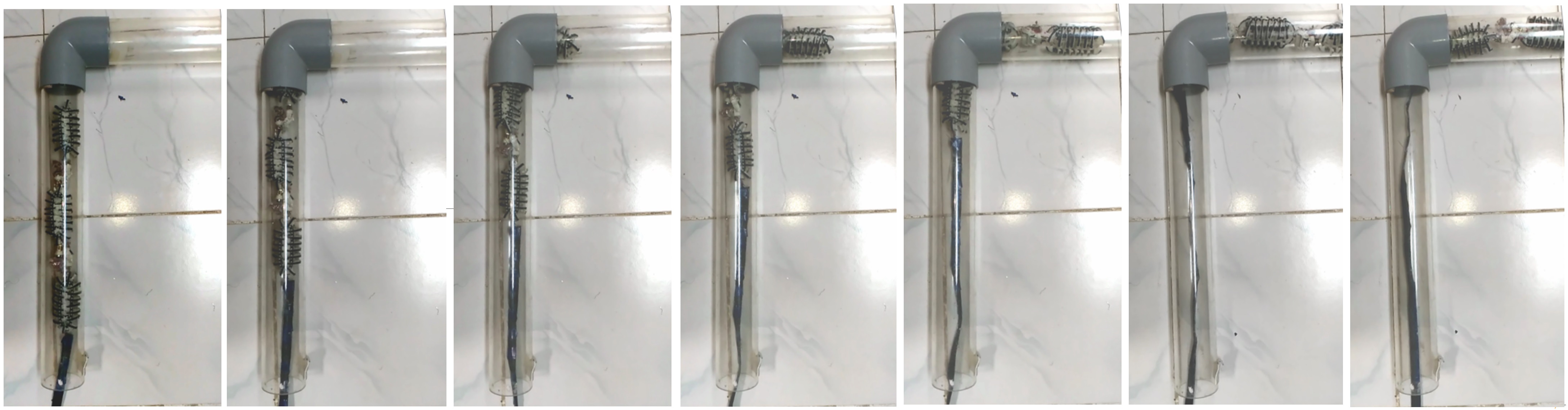}\label{fig:90 elbow}}

\subfloat[Experiment in $45^\circ$ bent elbow]{\includegraphics[width=1\textwidth,height=0.14\textheight]{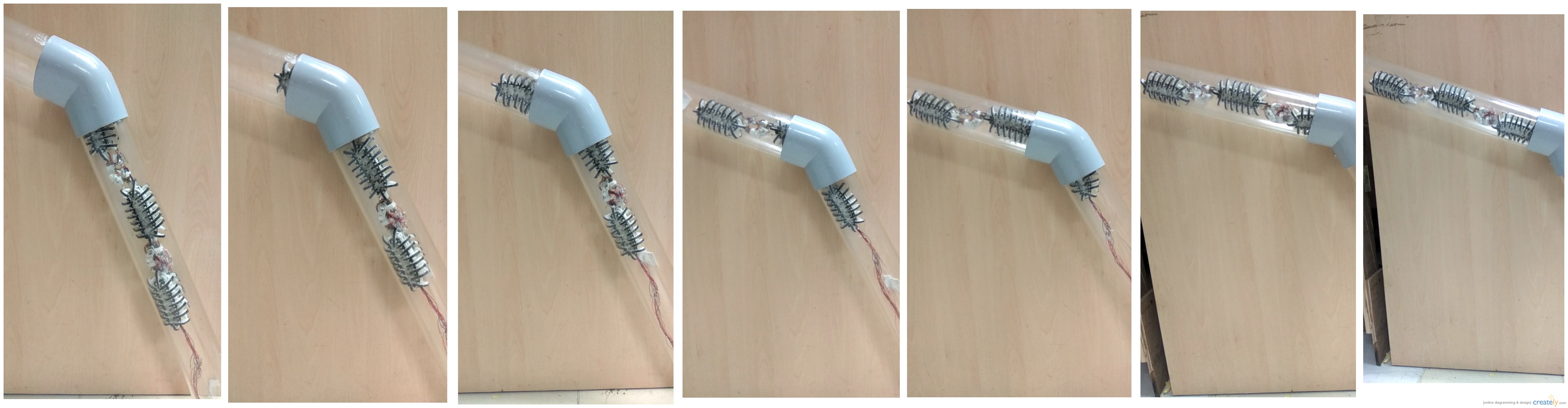}\label{fig:45 elbow}}\\    

\subfloat[Robot moves forward(A-B), then aligns itself along the bend by rolling along the pipe(C-D-E-F), then steers to overcome a 45$^\circ$ bend(G-H-I-J-K-L).]{\includegraphics[width=0.8\textwidth,height=0.24\textheight]{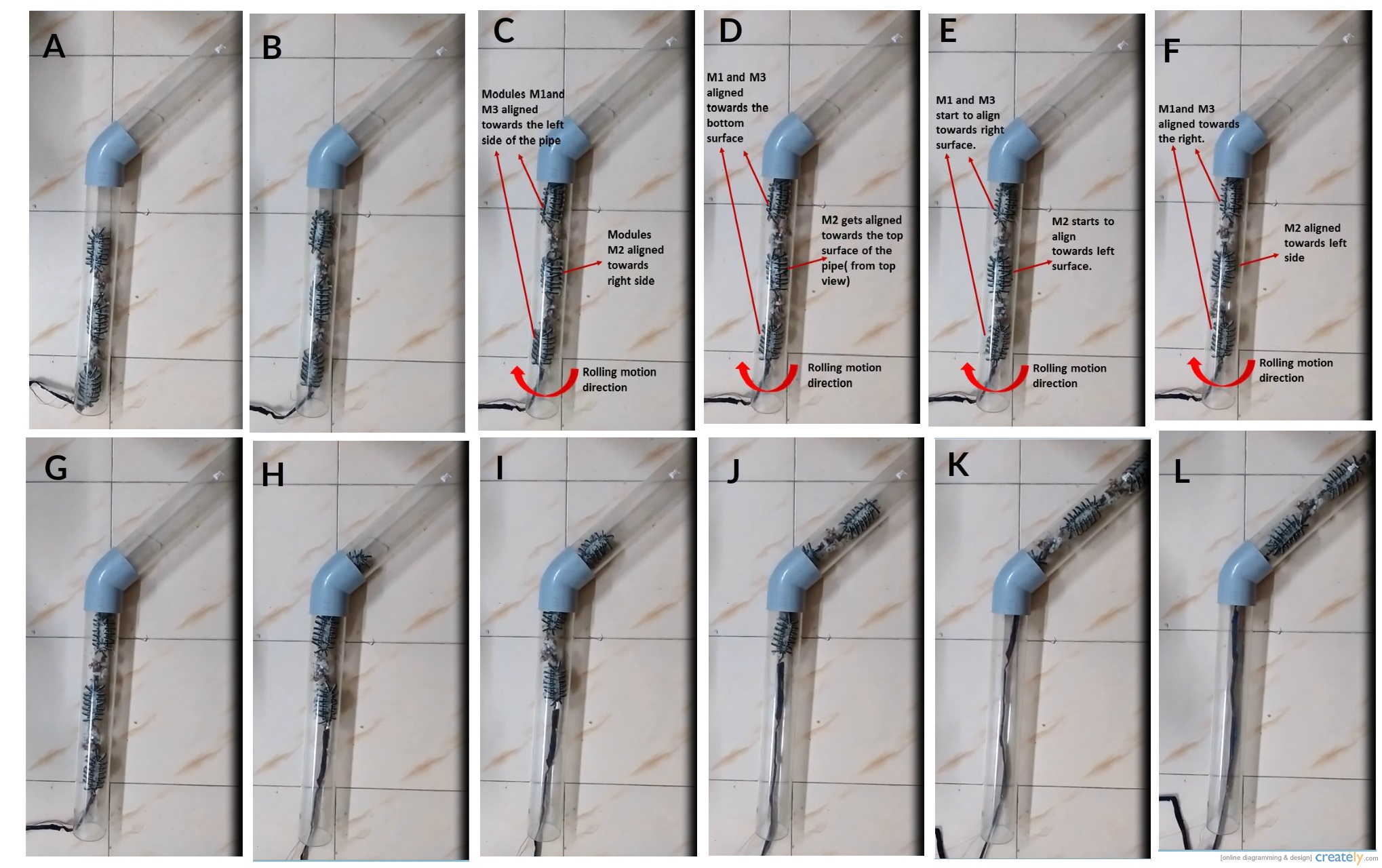}\label{fig:roll_turn}}\\    

\caption[Optional caption for list of figures 5-8]{Simulation and Experimental results demonstrating traversal in sharp turns.}    
\label{fig:90_degree}
\end{figure*}

\begin{align}
\begin{split}
 &\tau = k(\theta-\theta_{initial}),\\
 \text{where,} \quad &k: \text{spring stiffness}\\
                     &\theta: \text{current joint angle}\\
                     &\theta_{initial}:\text{initial joint angle (preloaded)}
\end{split}
\label{eqn:spring}
\end{align}

For vertical straight climbing in $\o$75mm pipes, the joint angle values obtained from equation(\ref{eqn:vert_theta}) are $\theta_1=115^\circ$ and $\theta_2=65^\circ$ and joint moment values are $\tau_{J_1}=0.2359$ Nm, $\tau_{J_2}=0.3683$ Nm, $\tau_{J_3}=0.2760$ Nm, $\tau_{J_4}=0.1310$ Nm, which result in the following values of spring stiffness.\\
$k_1$= 0.0096 Nm/deg,\quad
$k_2$ = 0.0056 Nm/deg,\\
$k_3$ = 0.0042 Nm/deg,\quad
$k_4$ = 0.0053 Nm/deg 

The values obtained above were successfully simulated to climb vertical pipes. However, the slippage was observed while steering in the bend direction. Therefore, a similar analysis was done for a $360^\circ$ circular pipe trajectory, as shown in Fig.\ref{fig:bend}, where each quadrant represents one of the possible configurations of a $90^\circ$ elbow(vertical to horizontal and horizontal to vertical, in both upward and downward directions) in a pipe network. 

The circular trajectory was discretized into $n$(360) steps and an optimization was carried out $\forall$ $l\in [1,n]$. The plot showing the optimal spring stiffness values obtained from optimization, $\forall l \in [1,n]$ is shown in Fig.\ref{fig:spring_stiff}. Here, it can be observed that the springs at joints $J_3$ and $J_4$ need to be more stiffer to balance the moments due to forces from the adjoining modules and links, which can also be illustrated from quasi-static equations. The pattern also shows that the stiffness values $k_1,k_2,k_3$ and $k_4$ are consistent for $\forall \phi \in [0,150^\circ]$ and therefore, the spring stiffness values were selected as the maximum within this range of $\phi$. The obtained values are listed below.\\
$k_1$ = 0.0262 Nm/deg, \quad
$k_2$ = 0.0170 Nm/deg,\\
$k_3$ = 0.0163 Nm/deg, \quad
$k_4$ = 0.0232 Nm/deg

These values were further successfully verified in simulation for $90^\circ$ bends in a $\textbf{\O}$75mm pipes, as shown in Fig \ref{fig:adams_90}. However, the sudden increase in the stiffness value $k_4$ at $\phi=320^\circ$, in Fig.\ref{fig:spring_stiff}, shows that the horizontal to vertically upward climbing via a $\phi=90^\circ$ elbow could not be addressed by these values.

In addition to above, the stiffness values obtained for vertical climbing were used to estimate ($\mu_{lim}$) from the formulated optimization in equation \ref{eqn:fric_opti}. The $F/N$ ratio obtained for modules $M1,M2$ and $M3$ are 0.5180, 0.4629 and 0.5054, respectively. Therefore,
\begin{align*}
\begin{split}
\mu_{lim}=max(0.5180,0.4629,05.5054)=0.5180 
\end{split}
\end{align*}

This implies that with the available driving motor saturation torques, the robot is able to successfully climb in vertical pipes with $\mu_{lim}<\mu<1$, with the designed springs. This $\mu_{lim}$ was further used to obtain spring stiffness values from Optimization(eqn.\ref{eqn:spring_opti}). Fig.\ref{fig:mu_lim} shows 2 curves for each $k_1,k_2,k_3,k_4$ representing the difference in the stiffness values obtained for a $\mu$ and $\mu_{lim}$. The difference indicates that the spring was designed for a particular $\mu$ and is able to work well with $\mu_{lim}$, by exploiting the driving motor torques well below its saturation limits. 

\begin{comment}
Moreover, the diameter of a pipe with a sharp 90 degree turn that can be climbed with these design parameters modules that can be climbed by the robot with the maximum joint limit angles of the folding modules, the diameter of the sharp 90 degree turn that can be tackled by the robot is that can be climbed without 
\end{comment}

\section{Experiments}
\label{experiment}
\begin{figure}[h!]
\centering
\subfloat[]{\includegraphics[width=0.14\textwidth,height=0.16\textheight,angle=-0] {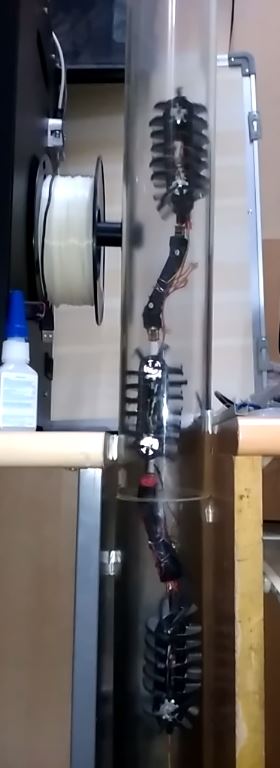}\label{fig:ADAMS_rolling}} 
\hspace{0cm}
\subfloat[ ]{\includegraphics[width=0.12\textwidth,height=0.16\textheight,angle=-0] {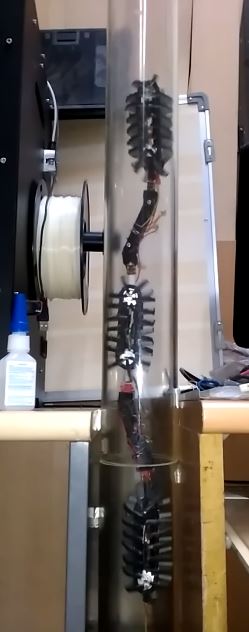}\label{fig:ADAMS_rolling}} 
\hspace{0cm}
\subfloat[ ]{\includegraphics[width=0.08\textwidth,height=0.16\textheight,angle=-0] {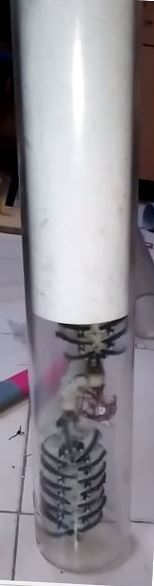}\label{fig:ADAMS_rolling}} 
\hspace{0cm}
\subfloat[ ]{\includegraphics[width=0.08\textwidth,height=0.16\textheight,angle=-0] {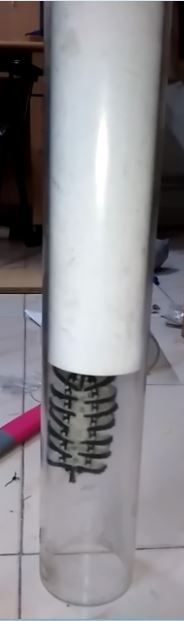}\label{fig:ADAMS_rolling}} 
\hspace{0cm}
\caption{(a),(b)) showing vertical climbing in acrylic pipe($\mu$=0.7); (c),(d) in glossy paper surface($\mu$=0.55)}
\label{fig:friction_vertical}
\end{figure}

Extensive experiments were conducted to validate the numerical and simulation results of the proposed design. The robot was manually controlled by an operator using DPDT switches. The initial prototype was developed with non-compliant modules with the optimal spring stiffness values. This could successfully climb vertical pipes, as demonstrated in Fig.\ref{fig:friction_vertical}. However, it could not negotiate \textbf{$\o$}75mm smooth elbows because of its diametric non-uniformity. This issue could be further solved by incorporating an active compliant joint in the module, which would successfully overcome sharp $45^\circ$ turns. However, this design was not realized in practice since sharp  $90^\circ$ turn still remained unaddressed due to the active compliance joint limit being \textbf{$40^\circ$}. This led to the design of a folding module with 2 active compliant joints, each with a joint angle limit of $35^\circ$ and all further experiments were performed using this prototype. To direct the robot to steer in the direction of a $45^\circ$ turn, the holonomic rolling motion was tested as demonstrated in Fig. \ref{fig:roll_turn}.
 
Subsequently, the vertical climbing was successfully tested in a pipe with a glossy paper on its surface, to validate the climbing ability in the limiting friction coefficient surface. This is depicted in Fig. \ref{fig:friction_vertical}.
\begin{comment}
\begin{figure*}[t]  
\centering    
\subfloat[]{\includegraphics[width=0.17\textwidth,height=0.16\textheight]{1.jpg}\label{fig:90_1}} 
\subfloat[]{\includegraphics[width=0.17\textwidth,height=0.16\textheight]{2.jpg}\label{fig:90_2}}
\subfloat[]
{\includegraphics[width=0.15\textwidth,height=0.16\textheight]{3.jpg}\label{fig:90_3}}  
\subfloat[]{\includegraphics[width=0.17\textwidth,height=0.16\textheight]{4.jpg}\label{fig:90_4}} 
\subfloat[]{\includegraphics[width=0.17\textwidth,height=0.16\textheight]{5.jpg}\label{fig:90_5}} 
\subfloat[]{\includegraphics[width=0.17\textwidth,height=0.16\textheight]{6.jpg}\label{fig:90_6}} \\
\subfloat[]{\includegraphics[width=0.17\textwidth,height=0.16\textheight]{8.jpg}\label{fig:90_7}} 
\subfloat[]{\includegraphics[width=0.17\textwidth,height=0.16\textheight]{9.jpg}\label{fig:90_7}} 
\subfloat[]{\includegraphics[width=0.17\textwidth,height=0.16\textheight]{10.jpg}\label{fig:90_7}} 
\subfloat[]{\includegraphics[width=0.17\textwidth,height=0.16\textheight]{11.jpg}\label{fig:90_7}} 
\subfloat[]{\includegraphics[width=0.17\textwidth,height=0.16\textheight]{12.jpg}\label{fig:90_7}} 
\subfloat[]{\includegraphics[width=0.17\textwidth,height=0.16\textheight]{13.jpg}\label{fig:90_7}} 

\caption[Optional caption for list of figures 5-8]{Experimental results demonstrating propagation in forward direction, followed by rolling and bending 45$^\circ$ bend to steer in the bends.}    
\label{fig: rolling_bending}
\end{figure*}
 \end{comment}

\section{Conclusion and Future Work}
In this paper, we discuss the design of a novel compliant Omnicrawler modular robot for in-pipeline climbing. The holonomic motion facilitates its alignment according to the direction of the bend, beforehand. With the given pipe environment, a set of optimal spring stiffness values were calculated to quasi-statically balance the robot in vertical as well as bend pipes. Afterwards, the limiting value of friction coefficient which could be climbed by the robot, without slippage was calculated.
This mechanism has the capability to climb in small diameter pipes, with smooth surface. However, the compliant active joint has 1 DOF, which restricts the compliance of the module in one direction. This requires robot to roll in such a way that the module could steer in that direction. Therefore, our future work would focus on modification of the design to address this issue. Furthermore, a torque control strategy for active compliance of each module would be implemented.


\begin{thebibliography}{99}

\bibitem{c1} Roslin, Nur Shahida, et al. "A review: hybrid locomotion of in-pipe inspection robot." Procedia Engineering 41 (2012): 1456-1462.

\bibitem{c2} Dertien, Edwin, et al. "Design of a robot for in-pipe inspection using omnidirectional wheels and active stabilization." Robotics and Automation (ICRA), 2014 IEEE International Conference on. IEEE, 2014.

\bibitem{c3}Debenest, Paulo, Michele Guarnieri, and Shigeo Hirose. "PipeTron series-Robots for pipe inspection." Applied Robotics for the Power Industry (CARPI), 2014 3rd International Conference on. IEEE, 2014.

\bibitem{c4}Roh, Se-gon, et al. "Modularized in-pipe robot capable of selective navigation inside of pipelines." Intelligent Robots and Systems, 2008. IROS 2008. IEEE/RSJ International Conference on. IEEE, 2008.

\bibitem{c5}Choi, Hyouk Ryeol, and Se-gon Roh. In-pipe robot with active steering capability for moving inside of pipelines. INTECH Open Access Publisher, 2007.

\bibitem{c6} Roh, Se-gon, and Hyouk Ryeol Choi. "Differential-drive in-pipe robot for moving inside urban gas pipelines." IEEE transactions on robotics 21.1 (2005): 1-17.

\bibitem{c7} Kakogawa, Atsushi, and Shugen Ma. "Design of a multilink-articulated wheeled inspection robot for winding pipelines: AIRo-II." Intelligent Robots and Systems (IROS), 2016 IEEE/RSJ International Conference on. IEEE, 2016.

\bibitem{c8}Hirose, Shigeo, et al. "Design of in-pipe inspection vehicles for/spl phi/25,/spl phi/50,/spl phi/150 pipes." Robotics and Automation, 1999. Proceedings. 1999 IEEE International Conference on. Vol. 3. IEEE, 1999.
 
\bibitem{c9} Kakogawa, Atsushi, Taiki Nishimura, and Shugen Ma. "Development of a screw drive in-pipe robot for passing through bent and branch pipes." Robotics (ISR), 2013 44th International Symposium on. IEEE, 2013.

	\bibitem{c10} Roh, Se-gon, et al. "In-pipe robot based on selective drive mechanism." International Journal of Control, Automation and Systems 7.1 (2009): 105-112.

\bibitem{c11}Kwon, Young-Sik, and Byung-Ju Yi. "Design and motion planning of a two-module collaborative indoor pipeline inspection robot." IEEE Transactions on Robotics 28.3 (2012): 681-696.

\bibitem{c12}Kakogawa, Atsushi, and Shugen Ma. "Design of an underactuated parallelogram crawler module for an in-pipe robot." Robotics and Biomimetics (ROBIO), 2013 IEEE International Conference on. IEEE, 2013.

\bibitem{c13}Tadakuma, Kenjiro, et al. "Crawler mechanism with circular section to realize a sideling motion." Proceeding of IEEE/RSJ International Conference on Intelligent Robots and Systems Acropolis Convention Center. Nice, 2008.

\bibitem{c14}Qian, Feifei, et al. "Walking and running on yielding and fluidizing ground." (2012): 345-352.


\bibitem{c15} Iagnemma, Karl, et al. "Planning and control algorithms for enhanced rough-terrain rover mobility." International Symposium on Artificial Intelligence, Robotics, and Automation in Space. Vol. 2. No. 2. 2001.


\bibitem{c16} Pratt, Gill A., and Matthew M. Williamson. "Series elastic actuators." Intelligent Robots and Systems 95.'Human Robot Interaction and Cooperative Robots', Proceedings. 1995 IEEE/RSJ International Conference on. Vol. 1. IEEE, 1995.
\end{thebibliography}
\end{document}